\documentclass[runningheads]{llncs}
\usepackage{graphicx}
\begin{document}
\title{Surgical Phase Recognition in Laparoscopic Cholecystectomy}

%
%
\author{Yunfan Li\inst{1} \and
Vinayak Shenoy\inst{1} \and Prateek Prasanna\inst{1} \and I.V. Ramakrishnan\inst{1} \and Haibin Ling\inst{1} \and Himanshu Gupta\inst{1} }
%
\authorrunning{Y. Li et al.}
%
\institute{Stony Brook University, Stony Brook NY 11790, USA 
}
\maketitle              
\begin{abstract}
Automatic recognition of surgical phases in surgical videos is a fundamental task in surgical workflow analysis. In this report, we propose a Transformer-based method that utilizes calibrated confidence scores for a 2-stage inference pipeline, which dynamically switches between a baseline model and a separately trained transition model depending on the calibrated confidence level. 
Our method outperforms the baseline model on the Cholec80 dataset, and can be applied to a variety of action segmentation methods.

\end{abstract}
\section{Introduction}
Surgical workflow analysis is an important field in robot-assisted surgery research, which has great potential in improving patient safety and achieving better surgery outcomes\cite{maier2017surgical}. In particular,  automatic segmentation of surgical steps is important for subsequent surgical training, intraoperative assistance and workflow optimization\cite{padoy2019machine,garrow2021machine}. 
\\\\
Early methods for surgical phase recognition utilized statistical models like conditional random field \cite{charriere2017real,quellec2014real} and hidden Markov models(HMMs) \cite{twinanda2016endonet,padoy2008line,dergachyova2016automatic}. However, these methods are limited in representation capacity and were unable to model complicated long-term temporal relations. Later, long short-term memory(LSTM) network \cite{hochreiter1997long} was introduced for long-term temporal dependency modeling. SV-RCNet \cite{jin2017sv} combined ResNet \cite{he2016deep} with LSTM in an end-to-end fashion to model the spatio-temporal relations between video frames. Czempiel et al. \cite{czempiel2020tecno} proposed a multi-stage TCN model named TeCNO, which explores long-term temporal relations from pre-computed spatial features.
\\\\
The introduction of Transformer \cite{vaswani2017attention} models saw a major shift in the paradigm of sequential modeling and significant improvement in various computer vision related tasks, including surgical phase recognition. Gao et al. \cite{gao2021trans} proposed a Transformer-based network which combines pre-computed spatial and temporal embeddings within a fixed temporal window to predict surgical phase labels. Czempiel et al. \cite{czempiel2021opera} introduced a novel attention regularization loss within the Transformer framework to encourage the model to focus on high-quality frames during training.
\\\\
Laparoscopic Cholecystectomy(LC) procedures in most cases follow a series of necessary steps \cite{ALES5766}. Based on this observation, we first devised an inference scheme which transitions between a set of 2-class classifiers based on previous and current predictions. The intuition is that in surgeries where the steps are strictly followed, there are only two possibilities as to which phase a certain frame can belong to. Additionally, we found that 2-class classifiers are better at distinguishing between two neighboring steps than a full-class classifier. We will evaluate this methodology and discuss its limitations.
\\\\
Subsequently, we devised another inference scheme based on confidence levels from the baseline model, the intuition being that when the baseline model is less confident about the prediction, we can switch to a 2-class model with higher accuracy.
In safety critical applications such as surgical phase recognition, classification models should not only be accurate, but also indicate when they are likely to be wrong  \cite{moon2020confidence}.
However, modern deep learning models like the Transformers usually suffer from miscalibration issues \cite{guo2017calibration}, therefore producing over-confident predictions, which limits the chance for human intervention or other alternative methods. Therefore, we propose a complementary scheme which addresses the problem of over-confident Transformer models using temperature scaling, an effective confidence calibration method, and achieves better results than the baseline model.

\section{Methodology}
\subsection{Dataset}
The Cholec80 dataset \cite{twinanda2016endonet} includes 80 videos of laparoscopic cholecystectomy procedures with a resolution of $1920\times 1080$ or $854\times 480$ pixels recorded at 25 frames-per-second(fps). The dataset defines 7 surgical stages for the procedure, and tool presence labels are provided for multi-task learning. Following previous works, all videos are subsampled to 1 fps, and all frames are resized to $250 \times 250$ pixels. For fair comparison, we use 40 videls for training, 8 videos for validation, and 32 videos for testing across all experiments.

\subsection{Transformer model}
Unlike a sequence-to-sequence model that processes input one at a time, a transformer model relies on self-attention mechanism to attend to the entire input sequence at once. The scoring functionality of the self-attention allows the model to focus on only the relevant parts of the input. 

\begin{figure}
\includegraphics[width=\textwidth]{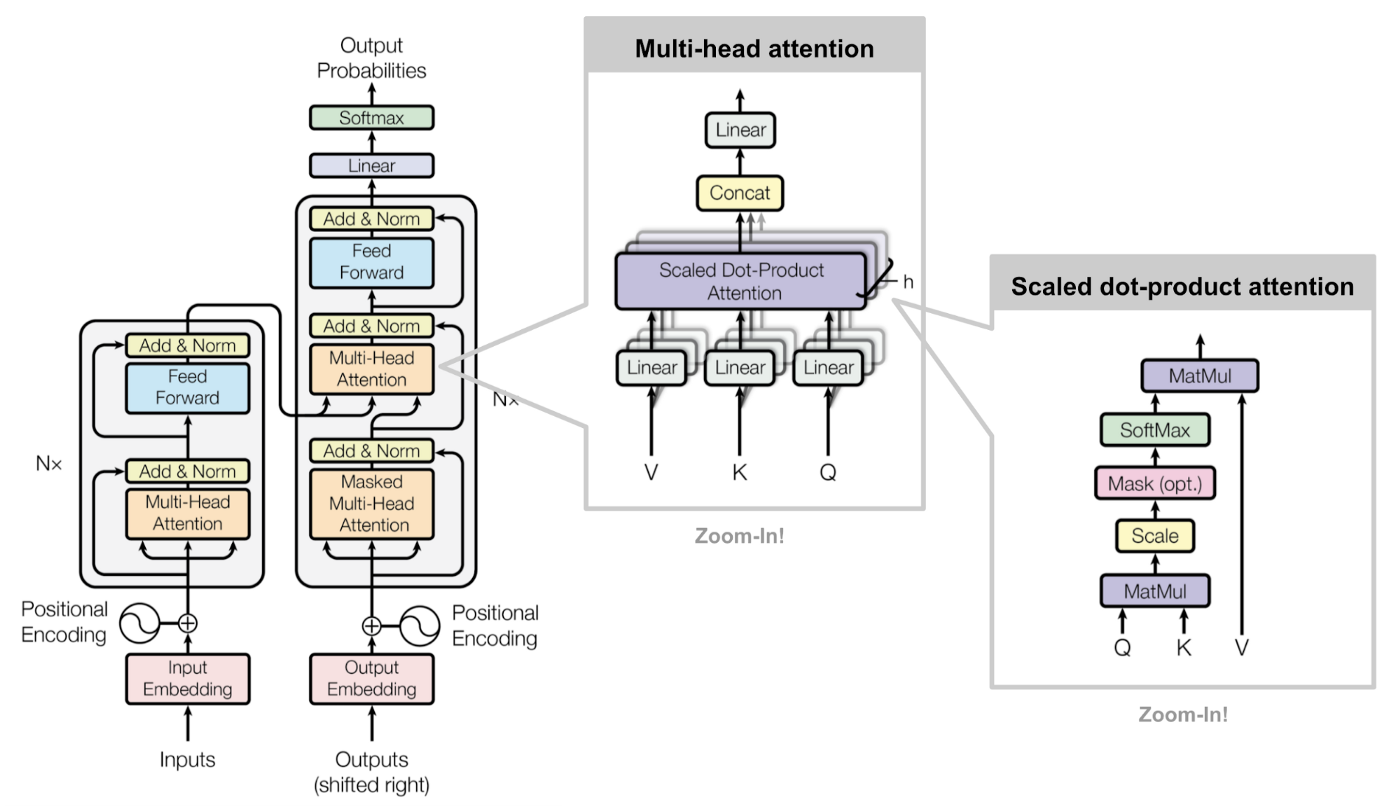}
\caption{Transformer Model} \label{fig1}
\end{figure}

The traditional transformer model (Fig. \ref{fig1}) is a stack of encoders and decoders. Each encoder if broken into two sub-layers: a multi-head self-attention mechanism and point-wise feed forward network. The output of an encoder layer is the input to the subsequent encoder layer in the stack. The decoder layer - in addition to the two sub-layers in the encoder - has another multi-head attention layer that performs attention over encoder output.

The attention function maps a query $Q$ with all the keys $K$. Its output is a weighted sum of all its value vectors $V$, where the weights are softmax scores computed using $Q$ and $K$. We can  think of the query as the search bar in youtube search that is mapped against a set of keys (video title, etc) associated with candidate videos in their database, then present the best matched values (videos). Each scaled dot-product attention unit in the multi-head attention learns a set of weights, $W_Q$, $W_K$, and $W_V$. The weights are multiplied with input, $x_t$ at timestep $t$ to produce query vector ($q_t = x_tW_Q$ ), key vector ($k_t = x_tW_K$) and value vector ($v_t = x_tW_V$). Each unit computes the attention of $Q$ with $K$ using:
$$
Attention(Q,K,V) = softmax(\frac{QK}{\sqrt{d_Q}})V
$$
where $Q$, $K$ and $V$ are matrices where $t$th row are $q_t$, $k_t$ and $v_t$ respectively. Each unit in the multi-head attention learns its own set of weights. The outputs are concatenated and fed to the feed-forward layers.

\subsection{Baseline model}
We adopt Trans-SVNet \cite{gao2021trans} as our baseline model, which achieved state-of-the-art results on the Cholec80 dataset. To be consistant with previous works, we first preprocess the dataset as specified in TMRNet \cite{jin2021temporal}.
Following the training pipeline of Trans-SVNet, we established a baseline result according to our dataset split. 

\subsection{Transition models}
Transition models are trained in the same way as the baseline model, except that we only train 2-class classifiers on two neighboring phases at a time. In total, this process yields 6 transition models. The intuition behind this approach is that 2-class predictors can better distinguish between two sequential phases than the 7-class model, as demonstrated in Table \ref{tab1}. 

\begin{table}[h]
\centering
\caption{2-class Transformer performance compared to baseline model.}\label{tab1}
\begin{tabular}{|c|c|}
\hline  
Model &  Accuracy (\%) \\
\hline
Trans-SVNet(baseline) &  87.44 \\
$Trans_{1-2}$ &  96.04 \\
$Trans_{2-3}$ & 95.19 \\
$Trans_{3-4}$ & 94.71\\
$Trans_{4-5}$ & 97.85\\
$Trans_{5-6}$ & 93.48\\
$Trans_{6-7}$ & 83.47\\
\hline
\end{tabular}
\end{table}

\subsection{Transition-based inference}
\subsubsection{Inference strategy}
Upon training the transition models, we designed our inference scheme as follows: (1) Initialize a buffer $B_N$ which stores the latest $N$ predictions and fill it with 1; (2) At each timestep, we look at the majority element in $B_N$, if the majority elements is $i ~(i<7)$, we use $Trans_{i\_(i+1)}$ for predicting the current label; if $i$ is 7, we use $Trans_{6\_7}$; (3) Append the current prediction to $B_N$ and pop out the oldest prediction from $B_N$. The buffer size $N$ is set to 100.
\subsubsection{Results and discussion}
\begin{figure}[h]
    \centering
    \caption{Example of cascading effect.}
    \includegraphics[width=0.5\textwidth]{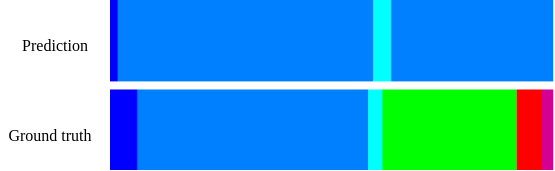}
    \label{fig:cascade}
\end{figure}

Table \ref{tab2} shows the results from transition-based inference strategy. As we can see, it does not show improvement upon the baseline model.
This strategy is good at avoiding noisy predictions from the baseline model, in which case the predicted phase labels tend to jump back and forth during transition periods. However, it also suffers from cascading effect as shown in Fig. \ref{fig:cascade}, which means a streak of incorrect predictions can lead to using the wrong transition model much further in the video, which causes bad overall results.

\begin{table}[h]
\centering
\caption{Results on different inference strategies.}\label{tab2}
\begin{tabular}{|c|c|}
\hline  
Model &  Accuracy (\%) \\
\hline
Trans-SVNet(baseline) &  87.44 \\
Transition-based & 86.92\\
Confidence-based w/o calibration & 65.64\\
Confidence-based w/ calibration & \textbf{88.02}\\
\hline
\end{tabular}
\end{table}

\subsection{Confidence-based inference}
Expanding on the previous strategy, we devised another strategy which relies on the confidence level output from the baseline model. Intuitively, when the output logits have a low value, it is less confident in its prediction than an otherwise high value. Based on this relative confidence level, we can set a threshold and switch to a transition model when the confidence of the model is below a desired level.

\subsubsection{Inference strategy}
We define our confidence-based inference strategy as follows: (1) Define a threshold $t_{conf}$ for the confidence score and the latest prediction $p_{last}=1$; (2) For each timestamp, use the baseline model to get a prediction $p_{base}$ and the corresponding confidence level $c_{base}$; if $c_{base}>t_{conf}$, update $p_{last}$ to $p_{base}$; if $c_{base}\leq t_{conf}$, we look at the value of $p_{last} = i ~(i<7)$, and use $Trans_{i\_(i+1)}$ to generate a substitute prediction $p_s$, then update $p_{last}$ to $p_s$; when $p_{last}$ is 7, we simply use $Trans_{6\_7}$.
\subsubsection{Confidence calibration}
While modern deep leaning models have dramatically improved neural network accuracy, they are also more prone to miscalibration(cite). This problem persists when applying our confidence-based inference strategy, as it heavily relies on reasonable confidence scores from the model.
To address this issue, we apply temperature scaling(cite) after training to achieve a better calibrated model. Temperature scaling uses a single scalar parameter $T>0$ for calibration. Given the logit vector $\textbf{z}$, the new confidence score is\\
\begin{equation}
\label{eq1}
    \hat{q} = \max_{k}\sigma_{SM}(\textbf{z}/T)^{(k)}
\end{equation}
$\sigma_{SM}$ is the Softmax(cite) function, whereas $k=\{1\ldots K\}$ is one of the $K$ classes. We use negative log-liklihood (NLL) \cite{guo2017calibration} and expected calibration error (ECE) \cite{guo2017calibration} as metrics for measuring the level of calibration. For both metrics, smaller values correspond to better calibration. The calibration results are shown in Table \ref{tab3}.

\begin{table}[!h]
\centering
\caption{Confidence calibration results.}\label{tab3}
\begin{tabular}{|c|c|c|}
\hline  
Model &  NLL & ECE \\
\hline
baseline &  0.576 & 0.215 \\
calibrated & 0.402 & 0.031\\
\hline
\end{tabular}
\end{table}

\subsubsection{Results and discussion}
Table \ref{tab2} shows the results of the confidence-based inference strategy. As we can see, without calibration, the strategy performs poorly due to over-confident logits from the baseline model. With temperature scaling, our strategy was able to outperform the baseline model, demonstrating the effectiveness of confidence calibration.

\section{Conclusion}
In this technical report, we explored incorporating transitional constraints into surgical phase recognition with Transformer models. Through experiments, we discovered that applying local transitional constraints can lead to catastrophic cascading effects, while in the meantime avoids noisy boundaries. Our proposed confidence-based inference strategy was able to leverage the superior performance of 2-class models and achieved better overall accuracy.





\newpage

\bibliographystyle{plain}
\bibliography{yunfan}

\begin{thebibliography}{10}

\bibitem{charriere2017real}
Katia Charri{\`e}re, Gw{\'e}nol{\'e} Quellec, Mathieu Lamard, David Martiano,
  Guy Cazuguel, Gouenou Coatrieux, and B{\'e}atrice Cochener.
\newblock Real-time analysis of cataract surgery videos using statistical
  models.
\newblock {\em Multimedia Tools and Applications}, 76(21):22473--22491, 2017.

\bibitem{czempiel2020tecno}
Tobias Czempiel, Magdalini Paschali, Matthias Keicher, Walter Simson, Hubertus
  Feussner, Seong~Tae Kim, and Nassir Navab.
\newblock Tecno: Surgical phase recognition with multi-stage temporal
  convolutional networks.
\newblock In {\em International conference on medical image computing and
  computer-assisted intervention}, pages 343--352. Springer, 2020.

\bibitem{czempiel2021opera}
Tobias Czempiel, Magdalini Paschali, Daniel Ostler, Seong~Tae Kim, Benjamin
  Busam, and Nassir Navab.
\newblock Opera: Attention-regularized transformers for surgical phase
  recognition.
\newblock In {\em International Conference on Medical Image Computing and
  Computer-Assisted Intervention}, pages 604--614. Springer, 2021.

\bibitem{dergachyova2016automatic}
Olga Dergachyova, David Bouget, Arnaud Huaulm{\'e}, Xavier Morandi, and Pierre
  Jannin.
\newblock Automatic data-driven real-time segmentation and recognition of
  surgical workflow.
\newblock {\em International journal of computer assisted radiology and
  surgery}, 11(6):1081--1089, 2016.

\bibitem{gao2021trans}
Xiaojie Gao, Yueming Jin, Yonghao Long, Qi~Dou, and Pheng-Ann Heng.
\newblock Trans-svnet: accurate phase recognition from surgical videos via
  hybrid embedding aggregation transformer.
\newblock In {\em International Conference on Medical Image Computing and
  Computer-Assisted Intervention}, pages 593--603. Springer, 2021.

\bibitem{garrow2021machine}
Carly~R Garrow, Karl-Friedrich Kowalewski, Linhong Li, Martin Wagner, Mona~W
  Schmidt, Sandy Engelhardt, Daniel~A Hashimoto, Hannes~G Kenngott, Sebastian
  Bodenstedt, Stefanie Speidel, et~al.
\newblock Machine learning for surgical phase recognition: a systematic review.
\newblock {\em Annals of surgery}, 273(4):684--693, 2021.

\bibitem{guo2017calibration}
Chuan Guo, Geoff Pleiss, Yu~Sun, and Kilian~Q Weinberger.
\newblock On calibration of modern neural networks.
\newblock In {\em International Conference on Machine Learning}, pages
  1321--1330. PMLR, 2017.

\bibitem{he2016deep}
Kaiming He, Xiangyu Zhang, Shaoqing Ren, and Jian Sun.
\newblock Deep residual learning for image recognition.
\newblock In {\em Proceedings of the IEEE conference on computer vision and
  pattern recognition}, pages 770--778, 2016.

\bibitem{hochreiter1997long}
Sepp Hochreiter and J{\"u}rgen Schmidhuber.
\newblock Long short-term memory.
\newblock {\em Neural computation}, 9(8):1735--1780, 1997.

\bibitem{jin2017sv}
Yueming Jin, Qi~Dou, Hao Chen, Lequan Yu, Jing Qin, Chi-Wing Fu, and Pheng-Ann
  Heng.
\newblock Sv-rcnet: workflow recognition from surgical videos using recurrent
  convolutional network.
\newblock {\em IEEE transactions on medical imaging}, 37(5):1114--1126, 2017.

\bibitem{jin2021temporal}
Yueming Jin, Yonghao Long, Cheng Chen, Zixu Zhao, Qi~Dou, and Pheng-Ann Heng.
\newblock Temporal memory relation network for workflow recognition from
  surgical video.
\newblock {\em IEEE Transactions on Medical Imaging}, 40(7):1911--1923, 2021.

\bibitem{maier2017surgical}
Lena Maier-Hein, Swaroop~S Vedula, Stefanie Speidel, Nassir Navab, Ron Kikinis,
  Adrian Park, Matthias Eisenmann, Hubertus Feussner, Germain Forestier,
  Stamatia Giannarou, et~al.
\newblock Surgical data science for next-generation interventions.
\newblock {\em Nature Biomedical Engineering}, 1(9):691--696, 2017.

\bibitem{ALES5766}
Arnab Majumder, Maria~S. Altieri, and L.~Michael Brunt.
\newblock How do i do it: laparoscopic cholecystectomy.
\newblock {\em Annals of Laparoscopic and Endoscopic Surgery}, 5(0), 2020.

\bibitem{moon2020confidence}
Jooyoung Moon, Jihyo Kim, Younghak Shin, and Sangheum Hwang.
\newblock Confidence-aware learning for deep neural networks.
\newblock In {\em international conference on machine learning}, pages
  7034--7044. PMLR, 2020.

\bibitem{padoy2019machine}
Nicolas Padoy.
\newblock Machine and deep learning for workflow recognition during surgery.
\newblock {\em Minimally Invasive Therapy \& Allied Technologies},
  28(2):82--90, 2019.

\bibitem{padoy2008line}
Nicolas Padoy, Tobias Blum, Hubertus Feussner, Marie-Odile Berger, and Nassir
  Navab.
\newblock On-line recognition of surgical activity for monitoring in the
  operating room.
\newblock In {\em AAAI}, pages 1718--1724, 2008.

\bibitem{quellec2014real}
Gw{\'e}nol{\'e} Quellec, Mathieu Lamard, B{\'e}atrice Cochener, and Guy
  Cazuguel.
\newblock Real-time segmentation and recognition of surgical tasks in cataract
  surgery videos.
\newblock {\em IEEE transactions on medical imaging}, 33(12):2352--2360, 2014.

\bibitem{twinanda2016endonet}
Andru~P Twinanda, Sherif Shehata, Didier Mutter, Jacques Marescaux, Michel
  De~Mathelin, and Nicolas Padoy.
\newblock Endonet: a deep architecture for recognition tasks on laparoscopic
  videos.
\newblock {\em IEEE transactions on medical imaging}, 36(1):86--97, 2016.

\bibitem{vaswani2017attention}
Ashish Vaswani, Noam Shazeer, Niki Parmar, Jakob Uszkoreit, Llion Jones,
  Aidan~N Gomez, {\L}ukasz Kaiser, and Illia Polosukhin.
\newblock Attention is all you need.
\newblock {\em Advances in neural information processing systems}, 30, 2017.

\end{thebibliography}

\end{document}